\documentclass{article}

\usepackage[preprint,nonatbib]{neurips_2022}

\usepackage[utf8]{inputenc} % allow utf-8 input
\usepackage[T1]{fontenc}    % use 8-bit T1 fonts
\usepackage{hyperref}       % hyperlinks
\usepackage{url}            % simple URL typesetting
\usepackage{booktabs}       % professional-quality tables
\usepackage{amsfonts}       % blackboard math symbols
\usepackage{nicefrac}       % compact symbols for 1/2, etc.
\usepackage{microtype}      % microtypography
\usepackage{xcolor}         % colors
\usepackage{amsmath}
\usepackage{amssymb}
\usepackage{graphicx}
\usepackage{caption}
\usepackage{subcaption}

\usepackage{biblatex} %Imports biblatex package
\usepackage{amsmath}

\DeclareMathOperator*{\argmin}{arg\,min}

\title{Active Learning with Tabular Language Models}

\author{%
  Martin Ringsquandl \\
  Siemens AG\\
  Munich, Germany \\
  \texttt{martin.ringsquandl@siemens.com}
  \And
  Aneta Koleva \\
  Siemens AG, LMU\\
  Munich, Germany \\
  \texttt{aneta.koleva@siemens.com}
}

\begin{document}

\maketitle

\begin{abstract}
  Despite recent advancements in tabular language model research, real-world applications are still challenging. In industry, there is an abundance of tables found in spreadsheets, but acquisition of substantial amounts of labels is expensive, since only experts can annotate the often highly technical and domain-specific tables. Active learning could potentially reduce labeling costs, however, so far there are no works related to active learning in conjunction with tabular language models. In this paper we investigate different acquisition functions in a real-world industrial tabular language model use case for sub-cell named entity recognition. Our results show that cell-level acquisition functions with built-in diversity can significantly reduce the labeling effort, while enforced table diversity is detrimental. We further see open fundamental questions concerning computational efficiency and the perspective of human annotators. 
\end{abstract}

\section{Introduction}
Recent tabular language models (TaLMs) which are based on pre-trained transformers and modality-adapted to tabular data have shown promising results on academic datasets \cite{dong2022}. However, using pre-trained transformer models for complex tasks in industry still requires fine-tuning with substantial amounts of labels. Although tabular data in spreadsheets is ubiquitous in industry, the application of TaLMs in real-world settings is still marginal. One reason is that acquiring labels for industrial spreadsheets, e.g. cell-level entity or column annotations, quickly becomes prohibitively expensive considering that often highly technical and domain specific language can only be annotated by a handful of experts. 

Settings with abundant unlabeled data and costly label acquisition are perfect matches for active learning (AL), which aims to minimize the costs of acquiring labeled data by maximizing model's performance with each new labeled instance. Deep neural networks, such as large-scale transformers, require batch-based acquisition functions instead of classical one-by-one instance selection, since these large models need to be trained for multiple epochs on batches of labeled data to effectively change the model's parameters for the next iteration of candidate acquisition. Otherwise training is inefficient and can lead to overfitting \cite{Ren2021}. A popular approach here is to have hybrid acquisition functions that weigh instance-wise uncertainty against some within-batch similarity (diversity) of all instances \cite{zhdanov2019}.

In the text data modality, batch typically means a set of sentences or documents. Even for a given token-level task such as named entity recognition (NER), we still want to select the most informative sentences (not single tokens), since all NER labels depend on the full sentence context. In AL terms, NER is a so-called multi-instance problem \cite{tomanek2009}. Transferring this to tabular data, a batch therefore should be a set of tables. Given a cell-level task, where each cell may have multiple labels, this is now a \textbf{nested} multi-instance problem. Since the instance is the full table, each table has multiple cells and cells have multiple labels. Until now, the table modality for AL has not been explored and it is unclear how to deal with such a novel AL problem.
In this paper, we study how well-known AL acquisition functions can be adopted to TaLMs to solve an industrial sub-cell NER task where named entities are mentioned in tables cells.

Our key contributions are the following:
\begin{itemize}
    \item We present an industrial NER use case for TaLMs in conjunction with AL.
    \item We outline novel problems that arise with AL when dealing with tabular data.
    \item We adopt well-known AL acquisition functions to tables and carry out an empirical evaluation on a real-world industrial dataset.
\end{itemize}

\section{Industrial Table NER Use Case}
\begin{figure}
    \centering
    \includegraphics[width=\textwidth]{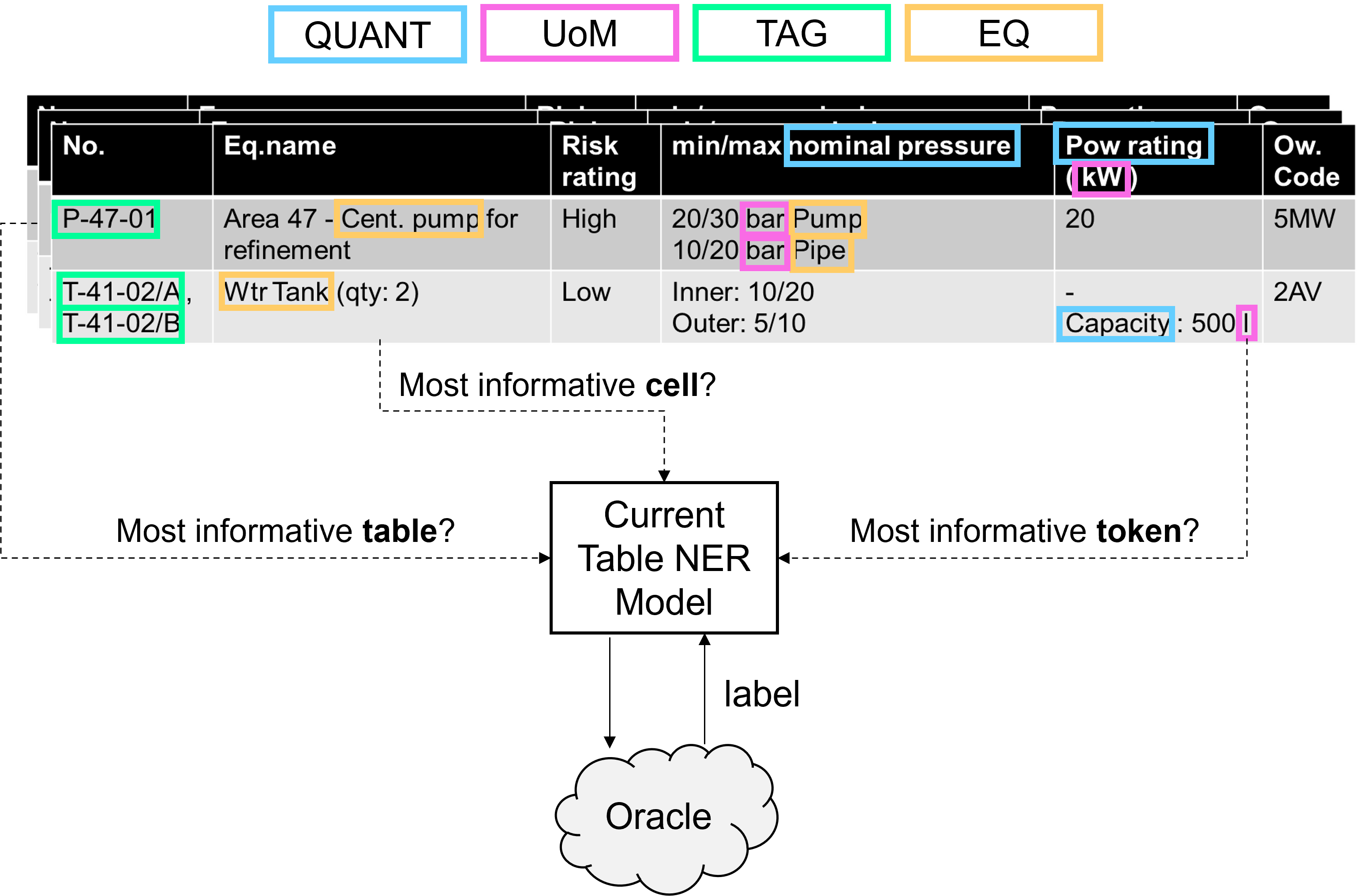}
    \caption{Overview of our industrial sub-cell NER problem and the active learning candidate acquisition.}
    \label{fig:example_table}
\end{figure}

Our use case originates from the need for information extraction to support data management in the process industry. Industrial plant operators have to maintain data about plant equipment, such as actuators, sensors, vessels, etc., processes, and ingredients among others. This data is used, for example, for process monitoring, maintenance, or other regulatory reasons. Typically, still today such data is collected and maintained by engineers in some form of spreadsheets. The spreadsheets are roughly organized in a tabular format, as shown in the example table in Figure~\ref{fig:example_table}. In these spreadsheets, each row typically represents information about one or multiple equipment instances. Some columns represent relevant physical properties of these equipment, while others are non-informative. However, the engineers do neither comply to a fixed schema, nor to unified spelling of equipment or properties. The NER task is to automatically extract relevant entities for creating structured specifications of the plant equipment. We phrase this problem as NER task (sub-cell NER) with the following types of entities. The type \textit{TAG} refers to a systematic identifier of an equipment. There are some conventions for generating equipment tags, but most plant operators customize them and some sheets do not contain identifiers at all. 
Type \textit{EQ} is for surface names of equipment types. The type \textit{QUANT} refers to the physical properties/quantities describing the functional specifications of equipment and the type \textit{UoM} stands for unit of measurement.

As these tables can get quite large, e.g. hundreds of rows, labeling all cells in a single table is already too time-consuming for human annotators. However, table cells can contain lots of similar, repetitive content, where it may be enough to provide small amount of labels to reach good performance. Therefore, we are interested in AL acquisition functions that would select the most informative instances to be labeled, but it is unclear on which level (token, cell, table) to select instances, as shown in the bottom of Figure~\ref{fig:example_table}.

\section{AL and TaLMs}
In this section we give a general introduction of AL and define our TaLM for sub-cell NER.

Given an unlabeled dataset $\mathcal{X}_U \sim \mathbb{P}$ sampled from some true unknown distribution $\mathbb{P}$. The objective of pool-based AL is to select candidates $\mathcal{X}_S \subset \mathcal{X}_U$ to be labeled with restricted budget $m \geq |\mathcal{X}_S|$ and $m \ll |\mathcal{X}_U|$ such that: 
\begin{equation}
\label{eq:AL}
    \argmin_{\mathcal{X}_S} \mathbb{E}_{x\sim \mathbb{P}} \left[ \ell(f_{\theta}(x), y) \right],
\end{equation}
where $y$ is the corresponding label of instance $x$, $\ell(\cdot)$ is a loss function and $f_{\theta}$ is a classification model trained on data ${\mathcal{X}_S}$ with revealed labels from an oracle.
Finding the optimal candidate set $\mathcal{X}_S$ w.r.t. \ref{eq:AL} is infeasible since in AL by definition we do not have access the true distribution nor any labels (except for a small seed set). However, based on principles such as uncertainty sampling AL research has proposed several heuristics to address this problem. These heuristics are often called query strategies or acquisition functions $f_{\text{acq}}(x_u; f_{\theta}): \mathcal{X}_U \rightarrow \mathbb{R}$ assigning a score to every instance based on the current classification model $f_{\theta}$.
Typically, AL is performed in an iterative fashion where candidates in the next iteration are sampled considering the model trained on the previous candidates.

In our case $f_{\theta}$ is a TaLM and $\mathcal{X}_U$ consists of a set of tables. We define a~table~as~a~tuple $T=(C,H)$, where $C = \{c_{1,1}, c_{1,2},  \dots c_{i,j} , \dots , c_{n,m} \}$ is the set of table body cells for $n$ rows and $m$ columns.~Every cell $c_{i,j} = \left( w_{c_{i,j},1}, w_{c_{i,j},2}, \dots, w_{c_{i,j},t} \right)$ is a sequence~of tokens of length $t$.~The table header $H = \{h_1, h_2, \dots, h_m\}$ is the set of corresponding $m$ column header cells, where $h_j =~\left(w_{{h_j},1}, w_{{h_j},2}, \dots , w_{{h_j},t}\right)$ is a sequence~of header tokens. 

Our TaLM consists of an encoder which will produce a table context-sensitive representation for each token: 
\begin{equation}
    \mathbf{w}_{h_{1},1}, \dots , \mathbf{w}_{h_{m},t}, \mathbf{w}_{c_{1,1},1}, \dots, \mathbf{w}_{c_{n,m},t} = \textsc{Enc}(T), 
\end{equation}
where $\textsc{Enc}$ is a table-biased transformer encoder using the row-column visibility matrix instead of vanilla attention and only within-cell positional encoding, refer to \cite{koleva2022} for a more detailed description of this architecture. Note that most TaLM architectures like TURL \cite{Deng2020} are not directly applicable here, since they already aggregate tokens within cells to get cell-level representations. This is why we resort to this  TaLM based on a pre-trained language model and modality adapt it to tables while fine-tuning for the sub-cell NER task. 

Each labeled cell has an NER-tag sequence: $(y_1, y_2, \dots, y_{t})$, where each $y_i \in \mathcal{Y}$. We use IO tags, thus $\mathcal{Y}$ is $\{\textit{O}\} \cup \{\textit{I-ENT}\}$, where $\textit{ENT} \in \{\textit{TAG},\textit{EQ},\textit{QUANT},\textit{UoM}\}$. Our TaLM decoder $\textsc{Dec}$ is a classification layer that projects each token representation into the label space, plus a Softmax activation to assign a normalized score for each class. 

\begin{equation}
    %\hat{y}_{1,1}, \dots, \hat{y}_{j,z} \dots \hat{y}_{m,t}, \hat{y}_{1,1,1}, \dots \hat{y}_{i,j,z}, \dots  \hat{y}_{n,m,t}
    \hat{\mathbf{y}}_{i,j,z} = f_{\theta}(T) = \textsc{Dec}(\textsc{Enc}(T)),
\end{equation}
where $\hat{\mathbf{y}}_{i,j,z}$ is the probability distribution of the labels of the $z$-th token in header or body cell $i,j$.

%In our sub-cell NER use case, we decided that human annotators would label on a cell-level and that labeling one cell has fixed cost on average no matter how many tokens in it.

%\subsection{Example Tasks}
%To keep things simple, we use entity linking as a task for TaLMs \cite{} to concretely show the implications of AL.

%Given a set of partially unlabeled tables $D = \{T_1, T_2, \dots, T_n\}$, each task instance $x_{k,i,j}$ corresponds to cell $c_{i,j}$ in table $T_k$ with corresponding true entity $y_{k,i,j}$. We can obtain a latent representation for this instance using the TaLM encoder with cell-wise pooling: $\mathbf{x}_{k,i,j} = \textsc{Cell-Pool}(\textsc{Enc}(T_k))_{i,j}$. On top of the TaLM encoder sits a classification head to model $P(y_{k,i,j} \mid \mathbf{x}_{k,i,j})$.

%By encoding every table, one can retrieve $P(y_{k,i,j} \mid \mathbf{x}_{k,i,j})$ for every unlabeled cell and apply any well-known query strategy, e.g. selecting instances with highest entropy $H(x_i)$.

\section{Table NER Acquisition Function}
As mentioned above, AL is primarily concerned with the choice of an acquisition function. This function assigns a score to each instance, thereby determining which new (unlabeled) instances to select for labeling such that they maximize the informativeness to the current model.

For single-instance tasks, such as text classification, the design of acquisition functions is straightforward - every instance has one label - so select the instance $x$ with the most informative label distribution $\hat{\mathbf{y}}$. For example, the popular entropy-based $f_{\text{acq}}^{\text{ent}}(x_U;f_{\theta}) = \mathbb{H}\left[\hat{\mathbf{y}} \mid x_U; f_{\theta} \right]$, simply ranks every $x_U$ based on the entropy $\mathbb{H}$ of the label distribution.

In multi-instance tasks such as NER there might be multiple labels (entity spans) in a single sentence instance. Therefore, we want to select the instance $x_U$ where the acquisition function considers the \textbf{joint} label distributions $\hat{\mathbf{y}}_1, \hat{\mathbf{y}}_2, \dots, \hat{\mathbf{y}}_t$.  In our table NER case, we have a \textbf{nested} multi-instance problem. Every table instance $T_k$ has multiple cells, for example body cells $c_{k,i,j}$, and every such cell has a sequence of label distributions $(\hat{\mathbf{y}}_{k,i,j,1}, \hat{\mathbf{y}}_{k,i,j,2}, \dots \hat{\mathbf{y}}_{k,i,j,t})$ associated with the $t$ tokens in the cell. 

%and every such cell has a NER-tag sequence $(y_{k,i,j,1}, y_{k,i,j,2} \dots y_{k,i,j,t})$ associated with the $t$ tokens in the cell. 

One could argue that in our table NER case, the acquisition function should select whole tables with most informative set of cells.
Unfortunately, it is not reasonable to assume that a human annotator would have the time to annotate every cell in a full table. Hence, since a single table would already break the annotation budget, we need an acquisition function on cell-level. One possibility is to do this hierarchically: first select a set of tables using $f_{\text{acq}}(T;f_{\theta})$ (e.g. by clustering table representations), then from this set of tables select a set of cells $f_{\text{acq}}(c_{i,j};f_{\theta})$ (e.g. by well-known multi-instance acquisition functions).

On the other hand, note that in our TaLM for NER $f_{\theta}(T)$ all cells $c_{i,j}$ in table $T$ are informed and therefore depend on each other. Based on this we argue that an acquisition function $f_{\text{acq}}(c_{i,j};f_{\theta})$ on cell-level is enough since it already captures the context of the full table $T$. 
%that make the simplifying assumption that cells within a table are independent 
In the following we adopt common acquisition functions to select batches of most informative cells. We consider different types of acquisition functions in our experiments - pure uncertainty-based, real-world diversity and batch diversity.

\paragraph{MNLP}
A popular acquisition function for uncertainty-sampling in multi-instance problems is the maximized normalized log-probability (MNLP) \cite{shen2017}. It considers the probability of the model's most likely label sequence to select new instances (here cells) $c_{i,j}$ with mean normalization to account for the bias towards longer sequences. Since the most likely label sequence in our TaLM is simply the sequence of every most likely token label we have:
\begin{align}
\label{eq:mnlp}
    f_{\text{acq}}^\text{MNLP}(c_{i,j}; f_{\theta}) = %\underset{\hat{\mathbf{y}}_{i,j,1},\dots,\hat{\mathbf{y}}_{i,j,t}}{\max} 
    \frac{1}{t} \sum_{z=1}^{t} \text{log} \text{ } y^*_{i,j,z}, %\text{ } \textit{p} (y_z \mid \mathbf{w}_{c_{i,j},1}, \dots,  \mathbf{w}_{c_{i,j},t})
\end{align}
where $y^*_{i,j,z}$ is the probability of the class with the maximum probability in the predicted label distribution. The sums of log-probabilities are negative, hence the candidates are selected in ascending order w.r.t. to~\ref{eq:mnlp}. MNLP has a major weakness in the batch-based AL setting. The sequences with highest uncertainty for the current model tend to be highly correlated. Therefore, when selecting a batch of candidates, there will be substantial similarity within the batch, providing little additional value. For tables this means that most candidates will likely come from the same table or even the same row/column.

\paragraph{MNLP+}
To make MNLP less prone to selection of similar cells, we introduce a simple procedure considering \textit{real-world diversity} in the tables. Instead of selecting the cells with the highest uncertainty from the set of all tables, we first sort all cells within every table w.r.t. to \ref{eq:mnlp}. Then we employ a round-robin selection across tables by taking the current table's best available candidate cell and moving to the next table until the budget is empty. In other words, this acquisition is a combination of cell uncertainty with a maximum table-diversity constraint.

\paragraph{BADGE}
%Others have suggested sampling from gradient embeddings instead of the regular model output which comes with implicit diversity 
As a batch-diverse acquisition function we consider Batch Active learning by Diverse Gradient Embeddings (BADGE) \cite{Ash2020}. Here, the authors propose to measure uncertainty in the magnitude of the model's last layer gradients. A diverse set of candidates is then selected using the $k$-\textsc{means}++ initialization to sample from the gradient embeddings. In our case, we take the mean of each cell's token gradient embeddings and feed this into $k$-\textsc{means}++ to get cell candidates:
\begin{align}
\label{eq:badge}
    f_{\text{acq}}^\text{BADGE}(c_{i,j}; f_{\theta}) = k\text{-}\textsc{means}\text{++}(\frac{1}{t}\sum_z \mathbf{g}_{i,j,z}),
\end{align}
where $\mathbf{g}_{i,j,z} = \frac{\partial}{\partial \theta^\text{out}}\ell_{\text{CE}}(\hat{\mathbf{y}}_{i,j,z}, y^*_{i,j,z}) $ is the gradient of the cross-entropy loss $\ell_{\text{CE}}$ of the $z$-th token label distribution and its predicted maximum label $y^*_{i,j,z}$ w.r.t. the decoder classification layer weights $\theta^\text{out}$. 

\paragraph{Rand}
As a simple baseline we also employ an acquisition function that selects candidate cells uniformly at random from the set of all cells in all available tables.

%Another particularity about tables is that columns tend to contain values of a consistent semantic type. E.g. "name" for person names. Should the AL selection criterion include something like diversity in table schema? Let the users annotate cells from the most different/diverse set of tables?

%The \LaTeX{} style file contains three optional arguments: \verb+final+, which
%creates a camera-ready copy, \verb+preprint+, which creates a preprint for
%submission to, e.g., arXiv, and \verb+nonatbib+, which will not load the
%\verb+natbib+ package for you in case of package clash.

%\paragraph{Preprint option}
%If you wish to post a preprint of your work online, e.g., on arXiv, using the
%NeurIPS style, please use the \verb+preprint+ option. This will create a
%nonanonymized version of your work with the text ``Preprint. Work in progress.''
%in the footer. This version may be distributed as you see fit. Please \textbf{do
%  not} use the \verb+final+ option, which should \textbf{only} be used for
%papers accepted to NeurIPS.

%At submission time, please omit the \verb+final+ and \verb+preprint+
%options. This will anonymize your submission and add line numbers to aid
%review. Please do \emph{not} refer to these line numbers in your paper as they
%will be removed during generation of camera-ready copies.

%The \verb+natbib+ package will be loaded for you by default.  Citations may be
%author/year or numeric, as long as you maintain internal consistency.  As to the
%format of the references themselves, any style is acceptable as long as it is
%used consistently.

\section{Experiments}
In this section we describe the experiments on our real-world industrial dataset.

\paragraph{Dataset}
The dataset contains spreadsheets from multiple industrial plants. The rows in the original sheets were downsampled, such that each resulting table had a maximum of 5 rows. Expert annotators were then asked to label each table cell-by-cell using the \url{prodi.gy} span-based NER annotation tool.
\begin{table}
  \centering
    \caption{Training tables statistics}
  \label{tab:training}
  \begin{tabular}{lc}
    \toprule
    Statistic  & Value \\
    \midrule
    \#tables  &  55 \\
    \#cells & 4,774      \\
    \#labels & 1,112  \\
    \midrule
    labels per cell & 0.23 \\
    \bottomrule
  \end{tabular}

\end{table}
After performing a random train-test split, we end up with 55 tables in the training set and 24 in the test set. The training set statistics are shown in Table~\ref{tab:training}. We can see that in total they contain $\sim$ 1,1k NER span labels. Dividing this by the number of cells in all tables, we observe that the vast majority $\sim$ 77\% of cells are only comprised of tokens with label \textit{O}. This is an extreme class-imbalance - typical for industry settings. Note that Figure~\ref{fig:example_table} is not representative of all tables, i.e. there are some with much less densely populated NER labels.

\paragraph{AL Setup}
For every AL acquisition function, a single experiment starts with a set of 100 labeled cell candidates $\mathcal{X}_L$ as seed to initially train our NER TaLM (iteration 0). Since tasks with high class-imbalance have strong dependence on this initial seed, we make sure that it is stratified w.r.t. the class distribution in the overall training set. This should mitigate the risk of completely missing minority classes in the initial seed.
\begin{figure}
     \centering
     \begin{subfigure}[b]{0.8\textwidth}
         \includegraphics[trim=1cm 1cm 2cm 2cm,clip=True ,width=1\textwidth]{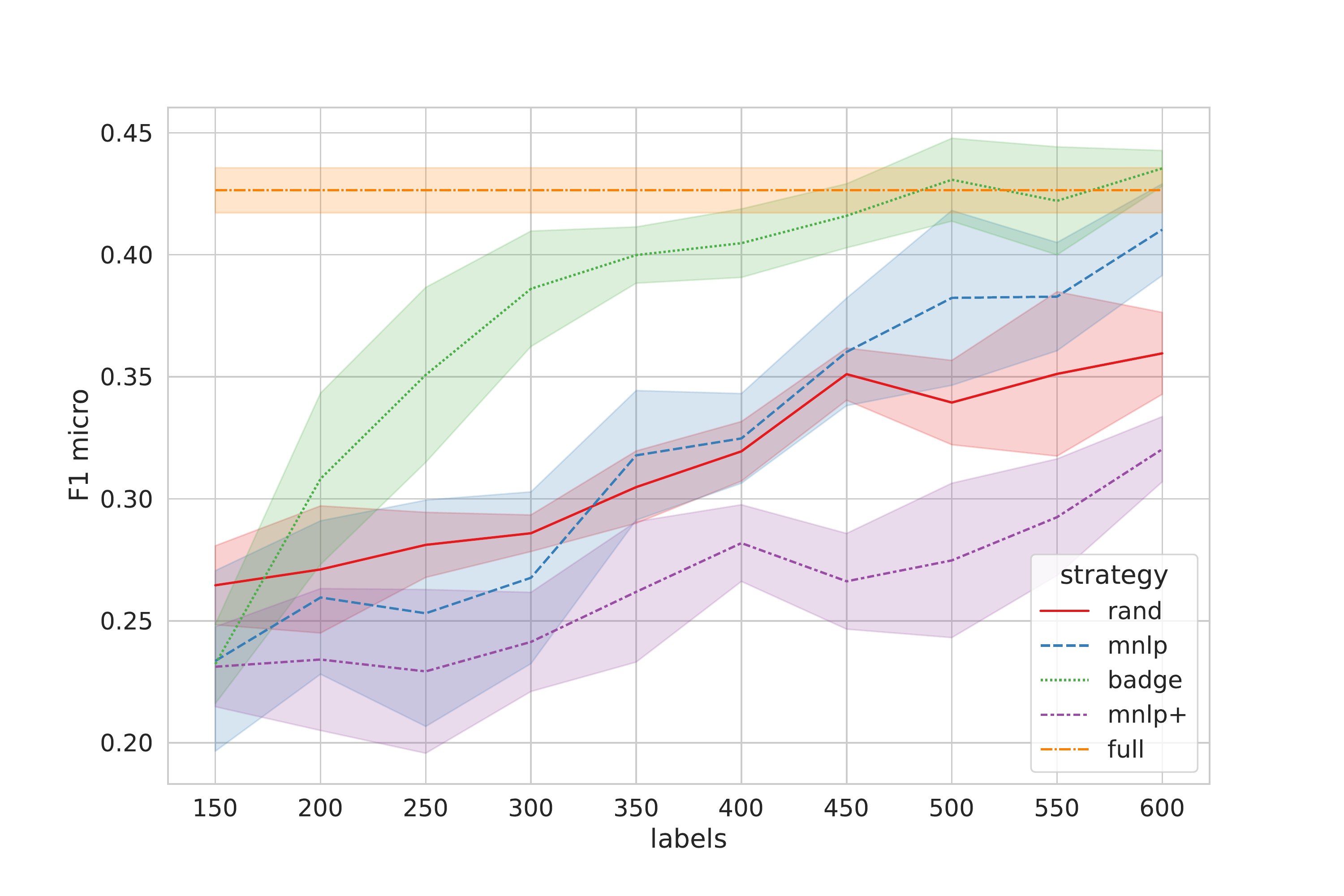}
        \caption{F1-score at each iteration averaged over 5 seeds}
        \label{fig:results}
     \end{subfigure}
     \hfill
     \begin{subfigure}[b]{0.8\textwidth}
         \includegraphics[trim=1cm 1cm 2cm 2cm,clip=True,width=1\textwidth]{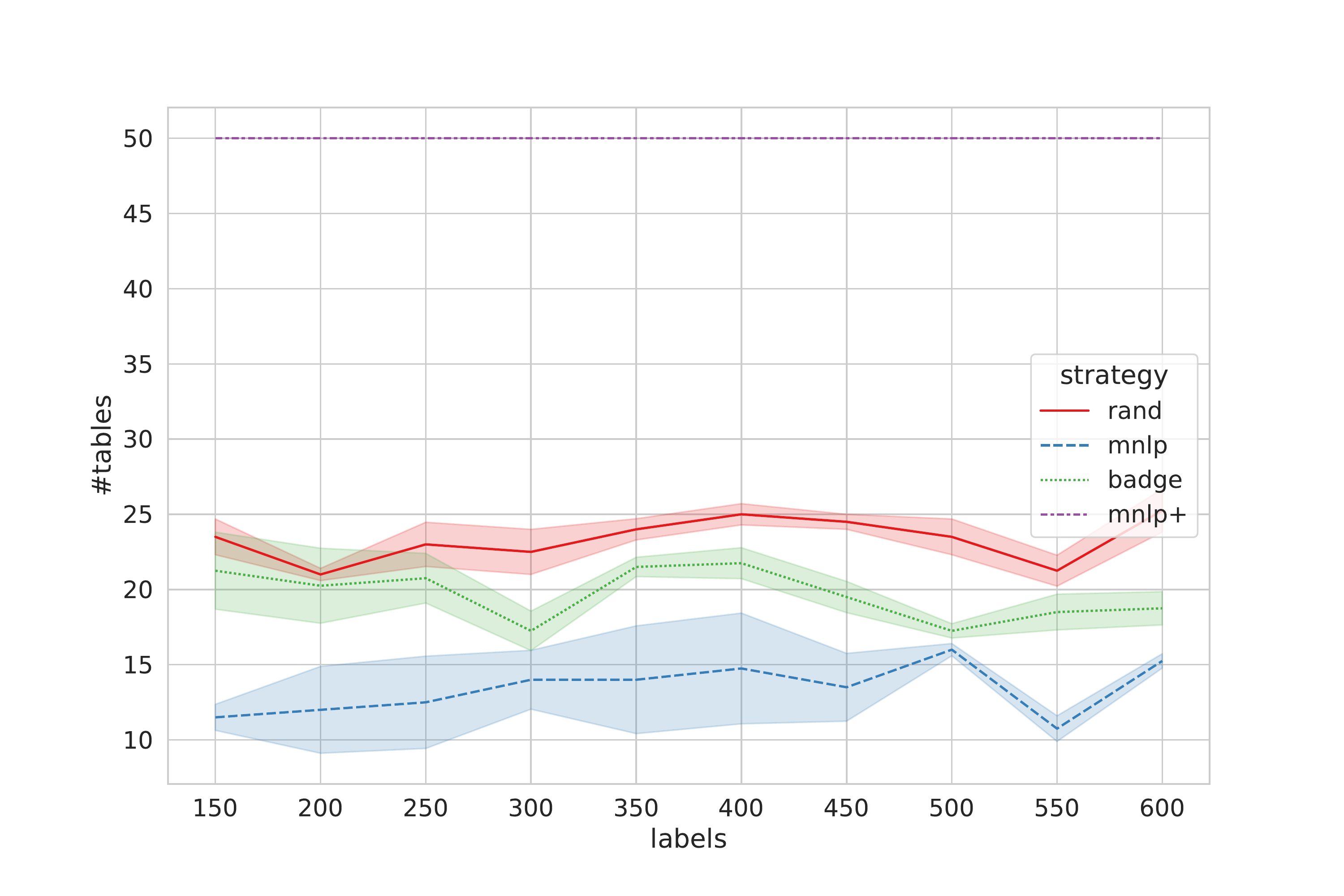}
         \caption{Number of tables sampled from at each iteration}
         \label{fig:num_tables}
     \end{subfigure}
     \caption{Results of different cell acquisition functions}
     \label{fig:both}
\end{figure}
After seed selection, the AL acquisition function selects a
batch $\mathcal{X}_S$ of 50 unlabeled instances (cells) from $\mathcal{X}_U$ that are added to $\mathcal{X}_L$ along with their true labels in each iteration. Following the suggested protocol from \cite{ein-dor2020}, we fine-tune in each iteration from scratch to avoid overfitting data from previous rounds. In each iteration we train for a maximum of 30 epochs and use a small portion of the test set as validation for early stopping. In each experiment, all acquisition functions start with the same initial seed. The reported results are the average over 5 different experiments, i.e., 5 different initial seeds. As pre-trained language model we use the 'bert-base-uncased' from huggingface.

After every experiment we evaluate the model's final performance against the hold-out test set. As a ceiling benchmark we also train our NER TaLM on the full training dataset. As performance metric we chose the micro-averaged F1-score.

\section{Results and Discussion}
The test set performance throughout iterations is shown in Figure~\ref{fig:results}. BADGE is the only acquisition function that clearly outperforms Rand and even exceeds full training performance. Interestingly, MNLP has only slight advantages over Rand after around 500 labels. MNLP+ is by far the worst meaning that the forced maximum table diversity is too extreme and detrimental to performance. We also notice higher variance for MNLP which suggests that it is more dependent on the initial seed labels, which is a known issue in these high-imbalance settings \cite{tomanek2009}. BADGE also seems to alleviate this problem to some degree looking more robust across different seeds. 

In terms of diversity of samples we show the number of different tables from which each acquisition function selects cells in each iteration in Figure~\ref{fig:num_tables}. %This metric seems to be correlated with the F1 scores, suggesting sampling from a wider range of tables is better. 
This metric shows that MNLP suffers from selecting similar cells from the same table with only slightly increasing number of tables. MNLP+ is by design selecting from as many tables as the budget allows. Random sampling only samples from roughly half the tables. This is due to an imbalance in table sizes - some large ones dominating the uniform sampling procedure. BADGE seems to strike a good balance between table diversity and cell uncertainty, sitting between MNLP and Rand in the number of tables.

Also note that the full training performance is lower compared to the original work \cite{koleva2022}, since we do not use data augmentation here. Although combining AL with data augmentation looks like a promising direction \cite{kim2021}.
%We notice the high variance between initial seed models. Coming up with better strategies to pick the initial seeds might be the most promising. 

%\subsection{Open Questions}
%\begin{itemize}
%    \item Diverse mini-batch AL: How to select instances for a mini-batch from a diverse set of tables?
%    \item For large tables: how to obtain appropriate sub-table(s)? Uncertainty metric may differ for a fixed instance $x_i$ given a subset of rows/columns.
%    \item 
%\end{itemize}

\paragraph{Computational Efficiency}
Apart from pure classification performance, we may also consider computational efficiency.
As an example, in a cell-level task, an acquisition function that retrieves a set of $k$ candidate cells from a single table versus $k$ cells which span over $k$ tables (one cell per table) will perform much fewer computations. Especially, for large tables we may want our acquisition to balance the informativeness of instances versus the number of tables that fit in a batch compute budget. Alternatively, we want the acquisition to select the most informative sub-tables rows (context), such that large tables will not impact the compute limits too much. Since TaLMs have a high memory footprint, a very table-diverse acquisition function may be much more expensive than others.

\paragraph{Human Annotators}
On the other hand, a table-diverse acquisition function may also impact human annotators. When presented with many different tables, the annotator probably needs more time per cell annotation, because every table's unique schema and context needs to be interpreted. This again raises the question if the full table needs to be shown to the user to make an informed labeling decision or a sub-table is sufficient.

%Depending on the level of the task, i.e. table- (table classification), column- (column type annotations), cell- (entity linking), or even sub-cell-level (sub-cell entity recognition) tasks

%Furthermore, for an NER-like task on table cells, we have to consider the change in data modality as well. In contrast to performing sentence-wise NER in text, where only a single sentence is input to the LM (e.g. BERT), now the full table is the input. 

%Table Slicing: Slicing out specific rows and columns of a table and comparing the informativeness scores
%Feeding sub-tables and continuously removing rows / columns with low attention values

\section{Related Work}
Pre-trained large-scale language models like BERT have been studied in AL setups for different tasks, e.g. text classification \cite{schroeder2022}.
While these models can already show good results when fine-tuning on small labeled datasets, in industry AL is still often necessary as argued in \cite{ein-dor2020}, where an empirical study of AL for BERT in text classification was carried out. They consider the challenging, practical scenario with high class-imbalance and small annotation budget. Their results show that batch-diverse acquisitions handle this scenario better than classical ones.

For NER deep conditional random field models have been studied with AL \cite{shen2017} with the focus on computational efficiency, due to expensive training. Further the \textit{missed-class effect} due to the AL's exploitative acquisition has been studied in NER \cite{tomanek2009}. The authors propose acquisitions for more informed seed instance selection that can alleviate this problem.

While all of these works are relevant to ours, it is non-trivial to transfer their findings to AL on tables using TaLMs.

\section{Conclusion}
In this paper we describe an industrial sub-cell NER use case for table language models and active learning. We define this novel nested multi-instance AL problem and adopt some well-known acquisition functions to select cell candidates from unlabeled tables. Our empirical results based on a real-world dataset show that these acquisition functions can be applied to TaLMs and outperform random selection. We find that balancing table diversity against label uncertainty is crucial and more advanced acquisition functions specifically for TaLMs and tables may be needed for other tasks. Since our work is the first to address this novel problem setting, there are still some open questions for future work regarding computational efficiency and impact on human annotators.

%\section*{References}
%\bibliographystyle{plainnat}
%\bibliography{library.bib}
\printbibliography %Prints bibliography

@InProceedings{schroeder2022,
  author =                {Christopher Schr{\"o}der and Andreas Niekler and Martin Potthast},
  booktitle =             {60th Annual Meeting of the Association for Computational Linguistics: Findings (ACL 2022)},
  ids =                   {potthast:2022d},
  month =                 may,
  publisher =             {Association for Computational Linguistics},
  site =                  {Dublin, Ireland},
  title =                 {{Revisiting Uncertainty-based Query Strategies for Active Learning with Transformers}},
  year =                  2022
}

@article{Ren2021,
author = {Ren, Pengzhen and Xiao, Yun and Chang, Xiaojun and Huang, Po-Yao and Li, Zhihui and Gupta, Brij B. and Chen, Xiaojiang and Wang, Xin},
title = {A Survey of Deep Active Learning},
year = {2021},
publisher = {ACM},
volume = {54},
number = {9},
journal = {ACM Comput. Surv.},
month = {oct},
articleno = {180}
}

@article{zhdanov2019,
  title={Diverse mini-batch active learning},
  author={Zhdanov, Fedor},
  journal={arXiv preprint arXiv:1901.05954},
  year={2019}
}

@inproceedings{ein-dor2020,
    title = "{A}ctive {L}earning for {BERT}: {A}n {E}mpirical {S}tudy",
    author = "Ein-Dor, Liat  and
      Halfon, Alon  and
      Gera, Ariel  and
      Shnarch, Eyal  and
      Dankin, Lena  and
      Choshen, Leshem  and
      Danilevsky, Marina  and
      Aharonov, Ranit  and
      Katz, Yoav  and
      Slonim, Noam",
    booktitle = "Proceedings of the 2020 Conference on Empirical Methods in Natural Language Processing (EMNLP)",
    month = nov,
    year = "2020",
    publisher = "Association for Computational Linguistics",
    pages = "7949--7962"
}

@inproceedings{dong2022,
author = {Dong, Haoyu and Cheng, Zhoujun and He, Xinyi and Zhou, Mengyu and Zhou, Anda and Zhou, Fan and Liu, Ao and Han, Shi and Zhang, Dongmei},
title = {Table Pre-training: A Survey on Model Architectures, Pre-training Objectives, and Downstream Tasks},
booktitle = {IJCAI'2022 SURVEY TRACK},
year = {2022},
month = {July}
}

@inproceedings{shen2017,
    title = "Deep Active Learning for Named Entity Recognition",
    author = "Shen, Yanyao  and
      Yun, Hyokun  and
      Lipton, Zachary  and
      Kronrod, Yakov  and
      Anandkumar, Animashree",
    booktitle = "Proceedings of the 2nd Workshop on Representation Learning for {NLP}",
    month = aug,
    year = "2017",
    address = "Vancouver, Canada",
    publisher = "Association for Computational Linguistics",
    pages = "252--256"
}

@article{Deng2020,
author = {Deng, Xiang and Sun, Huan and Lees, Alyssa and Wu, You and Yu, Cong},
title = {TURL: Table Understanding through Representation Learning},
year = {2020},
issue_date = {November 2020},
publisher = {VLDB Endowment},
volume = {14},
number = {3},
issn = {2150-8097},
journal = {Proc. VLDB Endow.},
month = {nov},
pages = {307–319},
numpages = {13}
}

@inproceedings{tomanek2009,
    title = "On Proper Unit Selection in Active Learning: Co-Selection Effects for Named Entity Recognition",
    author = {Tomanek, Katrin  and
      Laws, Florian  and
      Hahn, Udo  and
      Sch{\"u}tze, Hinrich},
    booktitle = "Proceedings of the {NAACL} {HLT} 2009 Workshop on Active Learning for Natural Language Processing",
    month = jun,
    year = "2009",
    publisher = "Association for Computational Linguistics",
    pages = "9--17",
}

@article{Ash2020,
  title={Deep Batch Active Learning by Diverse, Uncertain Gradient Lower Bounds},
  author={Jordan T. Ash and Chicheng Zhang and Akshay Krishnamurthy and John Langford and Alekh Agarwal},
  journal={ArXiv},
  year={2020},
  volume={abs/1906.03671}
}

@article{koleva2022,
   author = {Aneta Koleva and
            Martin Ringsquandl and
            Mark Buckley and
            Rakebul Hasan and
            Volker Tresp},
   title = {Named Entity Recognition in Industrial Tables using Tabular Language Models},
   journal={ArXiv},
   year={2022},
   volume={2209.14812}
}

@inproceedings{kim2021,
 author = {Kim, Yoon-Yeong and Song, Kyungwoo and Jang, JoonHo and Moon, Il-chul},
 booktitle = {Advances in Neural Information Processing Systems},
 editor = {M. Ranzato and A. Beygelzimer and Y. Dauphin and P.S. Liang and J. Wortman Vaughan},
 pages = {22919--22930},
 title = {LADA: Look-Ahead Data Acquisition via Augmentation for Deep Active Learning},
 volume = {34},
 year = {2021}
}

%%%%%%%%%%%%%%%%%%%%%%%%%%%%%%%%%%%%%%%%%%%%%%%%%%%%%%%%%%%%

%\appendix
%\section{Appendix}

%Optionally include extra information (complete proofs, additional experiments and plots) in the appendix.
%This section will often be part of the supplemental material.

\end{document}